\documentclass[letterpaper, 10 pt, conference]{ieeeconf}  

\IEEEoverridecommandlockouts                              

\usepackage{graphics} 
\usepackage{epsfig} 
\usepackage{mathptmx} 
\usepackage{amsmath} 
\usepackage{amssymb}  

\usepackage{arydshln}
\usepackage{nicematrix}
\usepackage{array}
\usepackage{cite}
\usepackage{nameref,hyperref}
\usepackage{caption}
\usepackage{subcaption}
\usepackage[utf8]{inputenc}
\usepackage{color,soul}
\usepackage[scr]{rsfso}
\usepackage{booktabs}
\usepackage{multirow}

\usepackage{here}

\title{\LARGE \bf
Frequency Domain Analysis of Nonlinear Series Elastic Actuator 

via Describing Function}

\author{Motohiro Hirao$^{1}$,  Burak Kurkcu$^{2}$, Alireza Ghanbarpour$^{3}$, and Masayoshi Tomizuka$^{4}$
\thanks{$^{1}$Motohiro Hirao is with NSK Ltd. Japan, and is a visiting fellow of the University of California, Berkeley, USA
        {\tt\small hirao@berkeley.edu}}%
\thanks{$^{2}$Burak Kurkcu is with the Department of Computer Engineering, Hacettepe University Turkey, and also is a Research Scholar at the University of California, Berkeley, USA
        {\tt\small bkurkcu@berkeley.edu}}%
\thanks{$^{3}$Alireza Ghanbarpour is with University of Tehran, IRAN, and is a   visiting PhD student of University of California at Berkeley, USA
        {\tt\small alireza.ghanbarpour@berkeley.edu}}%
\thanks{$^{4}$Masayoshi Tomizuka is with the Department of Mechanical Engineering,
University of California, Berkeley, MSC Lab, Etcheverry Hall, USA
        {\tt\small tomizuka@berkeley.edu}}%
}

\begin{document}

\maketitle
\thispagestyle{empty}
\pagestyle{empty}

\begin{abstract}

Nonlinear stiffness SEAs (NSEAs) inspired by biological muscles offer promise in achieving adaptable stiffness for assistive robots.
While assistive robots are often designed and compared based on torque capability and control bandwidth, NSEAs have not been systematically designed in the frequency domain due to their nonlinearity. The describing function, an analytical concept for nonlinear systems, offers a means to understand their behavior in the frequency domain. This paper introduces a frequency domain analysis of nonlinear series elastic actuators using the describing function method. This framework aims to equip researchers and engineers with tools for improved design and control in assistive robotics.

\end{abstract}

\section{INTRODUCTION}

In the era of rapidly advancing technology, human-robot interaction stands out as one of the most transformative technologies, especially in assistive robots. These robots, specifically designed to aid disabled individuals or enhance human capabilities, represent a fusion of engineering brilliance and human-centric design.
Assistive robots, especially those designed for direct physical interaction with humans, require a carefully designed actuator and controller to ensure soft and effective physical interaction.
This focus on design arises because both active and passive compliance is essential for adaptability and safety.

Series elastic actuators (SEAs) are commonly used for human-robot interaction because the elastic elements work as both force sensors and energy buffers~\cite {kong2011compact}. In SEA design, the elastic element is crucial since it decides the mechanical impedance and output force performance. Low stiffness introduces higher compliance but makes control bandwidth lower. Therefore, many conventional SEAs result in relatively higher stiffness design. 

To overcome the trade-off relationship caused by constant stiffness, variable stiffness actuators (VSAs) and nonlinear stiffness SEAs (NSEAs) have been explored in the literature. VSAs adjust their stiffness using either a secondary motor or a complex stiffness adjustment mechanism~\cite{li2019novel}. However, these actuators tend to be complicated and heavy. This complexity makes them difficult to control, especially in wearable assistive robots. 
On the other hand, NSEAs have a bio-inspired concept such as muscles~\cite{fahse2023dynamic} that has various stiffness according to the deflection of elastic elements and offer a promising solution to the limitation by simpler configuration than VSAs.
Many different NSEAs have been proposed and verified aiming to realize nonlinear stiffness with simple and compact configuration.

While these solutions present certain benefits, other designs also come with their set of challenges. NSEAs with rubber spring~\cite{austin2015control} or magnet~\cite{rafeedi2021design} realize compact and lightweight actuators. However, these methods introduce hysteresis and inaccuracy in stiffness characteristics. It makes control bandwidth narrower. ~\cite{zhu2021design} and ~\cite{bidgoly2016design} aimed for adjustable nonlinear stiffness, but their methods involved complex winding of pulley blocks or cam, leading to limitations from friction or speed dependency.
~\cite{thorson2011nonlinear} realized the nonlinear stiffness by combining coil spring and transmission, resulting in high accuracy to the low power-to-weight ratio due to the design.
In ~\cite{chen2019elbow}~\cite{qian2022design}, a novel series elastic actuator with nonlinear stiffness achieves a balance between low impedance, accurate force control, and simple configuration.

Assistive robots are commonly designed and classified based on output torque capability and control bandwidth ~\cite{sanchez2019compliant} since human movements are often analyzed in the frequency domain~\cite{winter2009biomechanics}.
Therefore, many conventional SEAs for practical applications have been designed based on requirements of maximum torque and control bandwidth. However, frequency domain analysis and design of NSEA is missing due to their nonlinear characteristics.

The describing function is a useful concept in the analysis of nonlinear systems~\cite{slotine1991applied}~\cite{gibson_describing_1962}. Originally developed as an extended frequency response analysis method, also it provides an approximate method to analyze the behavior of nonlinear systems~\cite{kurkccu2018disturbance}. Despite its approximate nature, this method has received significant attention and adoption~\cite{prawin2019damage}. The primary reasons for its popularity stem from its simplicity, making it a comprehensive method, and its robustness in providing a foundational basis for system design. In traditional system analyses, linear methods often do not capture the intricate behaviors of nonlinearities. The use of describing functions is not limited only to predicting limit cycles; they can also be applied in various analyses and designs.
To the best of our knowledge, there is no validation/derivation of the describing function analysis for the design and control of NSEAs.

In this paper, we delve into a frequency domain analysis of nonlinear series elastic actuators using the describing function method. This analytical framework offers a deeper understanding of the intricate dynamics of these actuators. Our approach aims to refine the tools for enhancing the design and control mechanisms in assistive robotics. As human-robot interactions become increasingly complex, methods such as ours may provide valuable insights for future innovations.

The remainder of this paper is organized as follows.
Section II discusses the elastic element design of a series elastic actuator in the case of linear stiffness and nonlinear stiffness. Section \ref{sec:dfnsee} derives the describing function of the nonlinear series elastic element (NSEE). Section \ref{sec:applicability} demonstrates the numerical verification results, and Section V concludes the work.

\section{ Series Elastic Element Design}\label{sec:seedes}

The elastic element must be carefully designed with their requirements since it highly affects human-robot interaction and force output performance of compliant actuators.
The actuator requirements for human-robot interaction and force output are evaluated in terms of torque amplitude and frequency domain characteristics ~\cite{sanchez2019compliant}.

In this section, we first provide a design procedure for linear stiffness elements based on the required output torque and saturation frequency, which represents the actuator's ability to oscillate at a full steady-state output force level~\cite{robinson2000design}. Next, we introduce a problem in nonlinear stiffness element design based on the required performance.

\subsection{Linear Stiffness Element Design} 

\begin{figure}[http]    
    \centering
    \includegraphics[width=205pt]{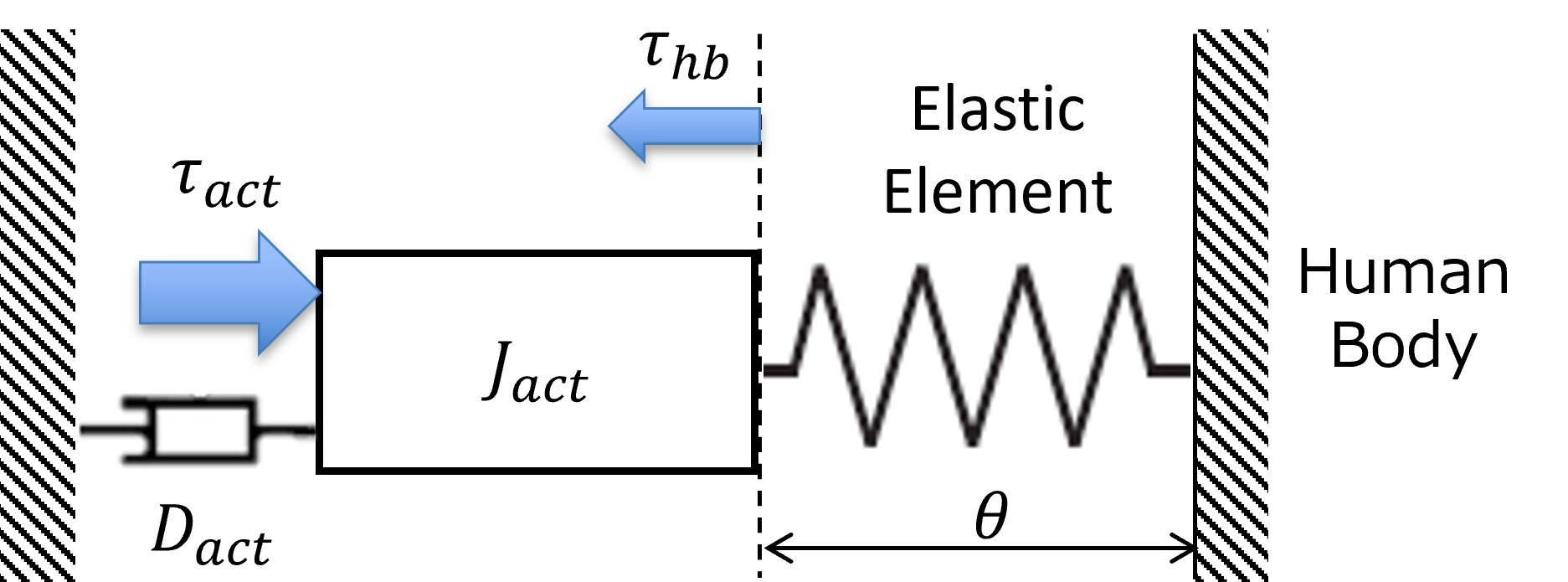}
    \caption{The Schematic of the simplified series elastic actuator. }
    \label{fig:sea}
\end{figure}

Fig.~\ref{fig:sea} presents simplified series elastic actuator including the equivalent inertia \textit{ $J_{act}$ }, and damping \textit{ $D_{act}$ }.
At this point, these values predominantly account for the inertia and damping of the motor as viewed through the transmission. Note that the human body serves as the reference for SEA. This modeling choice arises from the understanding that human body dynamics are generally slower than actuator dynamics~\cite{yu2015human} and typical human activities do not tend to destabilize the system state.

When the stiffness of the elastic element $k_{sea}$ is constant, the output torque of the elastic element $\tau_{ls}$ is defined as:

\begin{align}\label{eq:ksea}
\tau_{ls} = \tau_{hb} = k_{sea} \theta  
\end{align}

The dynamics of the actuator can be represented as:

\begin{align}\label{eq:seadyn}
J_{act} \Ddot{\theta} + D_{act} \Dot{\theta} + \tau_{hb} = \tau_{act}  
\end{align}

The transfer function from the torque generated by the motor to the output is given by:

\begin{align}\label{eq:seatf}
\frac{ \tau_{hb} (s) }{ \tau_{act} (s) } = \frac{ k_{sea} \theta (s) }{ \tau_{act} (s) } = \frac{ k_{sea} } {J_{act} s^2 + D_{act} s + k_{sea} }  
\end{align}

The saturation frequency~\cite{dos2017design} where the gain of the system begins to decrease could be expressed by \eqref{eq:seatf} as:
\begin{align}\label{eq:wsat}
\omega_{sat} = \sqrt{ \frac{k_{sea}}{J_{sea}} } 
\end{align}

From \eqref{eq:ksea} and \eqref{eq:wsat}, we can decide the stiffness $k_{sea}$ by requirement for maximum torque $ \tau_{max} $ and saturation frequency $ \omega_{sat} $, maximum deflection $ \theta_{max} $ as:

\begin{align}\label{eq:kdes}
k_{sea} = \omega_{sat}^2 J_{sea} = \frac{ \tau_{max} }{ \theta_{max} }  
\end{align}

\subsection{Nonlinear Stiffness Element Design} 
\begin{figure}[http]
    \centering
    \includegraphics[width=205pt]{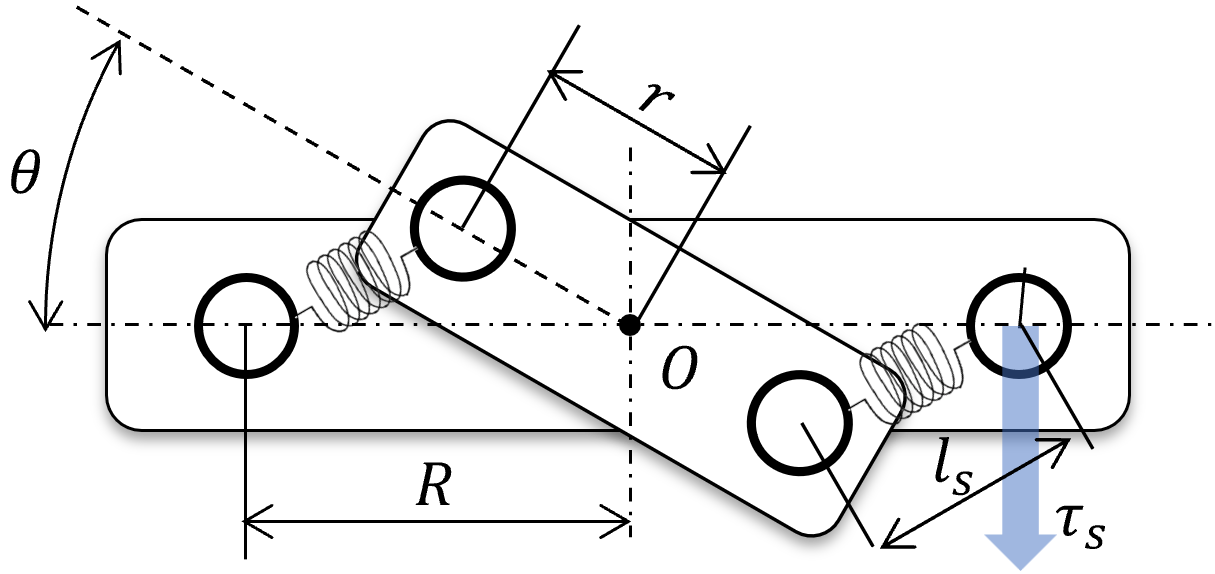}
    \caption{The Schematic of Nonlinear Stiffness Elastic Element. }
    \label{fig:nsee}
\end{figure}

A schematic representation of the NSEE is described in Fig.~\ref{fig:nsee}. In this design, two bars rotate coaxially around point \textit{O} and are coupled by tension springs. While one bar is driven by the motor via the transmission mechanism, the other is linked to the external load.

The spring length  $l_s$  is given by:

\begin{align}\label{eq:ls}
l_s = \sqrt{R^2 + r^2 - 2Rr \cos\theta}
\end{align}

Assuming the spring's rest length is $R-r$ and using Hooke's law, we get the tension force $F_s$  of each spring, factoring in spring stiffness $k_s$ and length: 

\begin{align}\label{eq:Fs}
F_s = k_s \{ l_s - (R-r) \}
\end{align}

The output torque of the NSEE $\tau_{ns}$ can be computed by:

\small
\begin{align}\label{eq:tns}
\tau_{ns} = nF_s \frac{Rr \sin\theta}{l_s} = nk_sRr (1 - \frac{(R-r)}{\sqrt{R^2 + r^2 - 2Rr \cos\theta} } ) \sin\theta
\end{align}
\normalsize

where \textit{n} corresponds to the number of springs.

The stiffness of the NSEE $k_{ns}$ varries with $\theta$ as:

\begin{align}\label{eq:kns}
k_{ns} ( \theta ) = \frac{ d \tau_{ns} (\theta) }{ d \theta } 
\end{align}

This varied stiffness with respect to the angle introduces better human-robot interaction performance compared to linear stiffness elements~\cite{qian2022design}.
Although control bandwidth remains a pivotal requirement for actuators, the design and analysis of such nonlinear stiffness actuators in the frequency domain are challenging since the saturation frequency depends on input torque bandwidth. Nonetheless, the describing function, an extended frequency response analysis method, can be employed for approximate nonlinear behavior analysis~\cite{slotine1991applied}.
The subsequent section of this paper presents the derivation and validation of the describing function for the NSEE.

\section{Describing Function of NSEE} \label{sec:dfnsee}

The describing function method offers a unique lens for analyzing nonlinear systems. In this section, we employ this technique to elucidate the frequency domain characteristics of nonlinear series elastic actuators, paving the way for more informed robotic design decisions.

Let $ \alpha = \frac{2 ( R - r )^2 }{ R r } $, $ \beta = R r $, \eqref{eq:tns} can be expressed by using Maclaurin expansion under certain conditions as:

\begin{align}\label{eq:mac}
\tau_{ns} = n k_s \beta \frac {1} { \alpha + \theta^2 } \theta^3
\end{align}

Suppose that the output of the nonlinear spring to the sinusoidal excitation  $\theta$ is in the form of $\theta = A \sin ( \omega t )$, $\tau_{ns}$ can be represented by the following Fourier series:

\begin{align}\label{eq:fe}
\tau_{ns} (t) = \frac{a_0}{2} + \sum_{n=1}^{\infty} (a_n \cos \omega t + b_n \sin \omega t)
\end{align}

We assume that the nonlinear system is symmetrical, given that $\tau_{ns}$ is an odd function.
Therefore, $a_0 = 0$,  and our focus is solely on the fundamental components $a_1$ and $b_1$~\cite{slotine1991applied},

\begin{align} \label{eq:fe2}
\tau_{ns} (t) = a_1 \cos \omega t + b_1 \sin \omega t
\end{align}
where the Fourier coefficients $a_1$ and $b_1$ can be determined by integrating over a complete cycle, which results in

\begin{align} \label{eq:a1}
a_1 &= \frac{1}{\pi} \int_{-\pi}^{\pi} \tau_{ns} (t) \cos \omega t d \omega t \nonumber
\\ &= \frac{ n k_s \beta }{\pi} \int_{ -\pi }^{ \pi } \frac{1}{ \alpha + ( A \sin \omega t )^2 } (A \sin \omega t )^3 \cos \omega t d \omega t
\end{align}    

\begin{align} \label{eq:b1}
b_1 &= \frac{1}{\pi} \int_{-\pi}^{\pi} \tau_{ns} (t) \sin \omega t d \omega t \nonumber
\\ &= \frac{ n k_s \beta }{\pi} \int_{ -\pi }^{ \pi }  \{ A \sin^2 \omega t - \frac{ \alpha }{ A } + \frac{ \alpha^2  (  b_{11} + b_{12} ) }{ 2 A \sqrt{ \alpha + A^2 }} \} d \omega t 
\end{align}

where $b_{11} = \frac{1}{ \sqrt{ \alpha + A^2 } + A \cos  \omega t } $  $ b_{12} = \frac{1}{ \sqrt{ \alpha + A^2 } - A \cos  \omega t } $.

Let $ x = \sin \omega t$, $ dx = \cos \omega t d \omega t$,

\begin{align} \label{eq:a1_3}
a_1 = \frac{ n k_s \beta }{\pi} \int_{ \sin -\pi }^{ \sin \pi } \frac{1}{ \alpha + ( Ax )^2 } (Ax)^3 d x = 0
\end{align}

Let $ z = e^{i \omega t } $, $ \cos \omega t = \frac{1}{2} ( z + \frac{1}{z} )$, $ d \omega t = -  \frac{i}{z} dz $,

\begin{align}\label{eq:b11}
 \int_{-\pi}^{\pi} b_{11} d \omega t 
 &= - \frac{1}{A} \oint_{ e^{ - i \pi } }^{ e^{ i \pi } } \frac{ 2 i }{ z^2 + 2 z \sqrt{ \frac{ \alpha }{ A^2 } + 1 } + 1 } d z \nonumber
\\ &= - \frac{1}{A} \oint_{ e^{ - i \pi } }^{ e^{ i \pi } } f_1(z) d z 
\end{align}    

\begin{align}\label{eq:b12}
\int_{-\pi}^{\pi} b_{12} d \omega t &= \frac{1}{A} \oint_{ e^{ - i \pi } }^{ e^{ i \pi } } \frac{ 2 i }{ z^2 - 2 z \sqrt{ \frac{ \alpha }{ A^2 } + 1 } + 1 } d z \nonumber
\\ &= \frac{1}{A} \oint_{ e^{ - i \pi } }^{ e^{ i \pi } } f_2(z) d z 
\end{align}    

Note that $f_1$ and $f_2$ have a single pole within the interval of integration.  

According to the residue theorem\cite[Section 5]{esl2012complex} ,

\begin{align} \label{eq:f1}
\oint_{ e^{ - i \pi } }^{ e^{ i \pi } } f_1(z) d z &= 2 \pi i Res ( f_1, - \sqrt{ \frac{ \alpha }{ A^2 } + 1 } + \sqrt{ \frac{ \alpha }{ A^2 } } ) \nonumber
\\ &= 2 \pi i Res ( f_1, p_1 )
\\ &= \lim_{z \to p_1 } ( z - p_1 ) f_1 ( z ) = - \frac{ 2 \pi A }{ \sqrt{ \alpha } } \nonumber
\end{align}   

\begin{align}\label{eq:f2}
\oint_{ e^{ - i \pi } }^{ e^{ i \pi } } f_2(z) d z &= 2 \pi i Res ( f_2, \sqrt{ \frac{ \alpha }{ A^2 } + 1 } - \sqrt{ \frac{ \alpha }{ A^2 } } ) \nonumber
\\ &= 2 \pi i Res ( f_2, p_2 )
\\ &= \lim_{z \to p_2 } ( z - p_2 ) f_2 ( z ) = \frac{ 2 \pi A }{ \sqrt{ \alpha } } \nonumber
\end{align}    

From \eqref{eq:fe2} -  \eqref{eq:f2}, $\tau_{ns}$ could be represented by using describing function $N_{ \tau } (A) $ as:
\begin{align} \label{eq:df}
\tau_{ns} &= n k_s \beta \{ A + \frac{ 2 \alpha }{ A } ( \frac{ 1 }{ \sqrt{ 1 + \frac{ A^2 }{ \alpha } } } - 1 ) \} \sin \omega t \nonumber
\\ &= n k_s \beta \{ 1 + \frac{ 2 \alpha }{ A^2 } ( \frac{ 1 }{ \sqrt{ 1 + \frac{ A^2 }{ \alpha } } } - 1 ) \} \theta 
= N_{ \tau } (A) \theta
\end{align}

\begin{figure}[http]
    \centering
    \includegraphics[width=205pt]{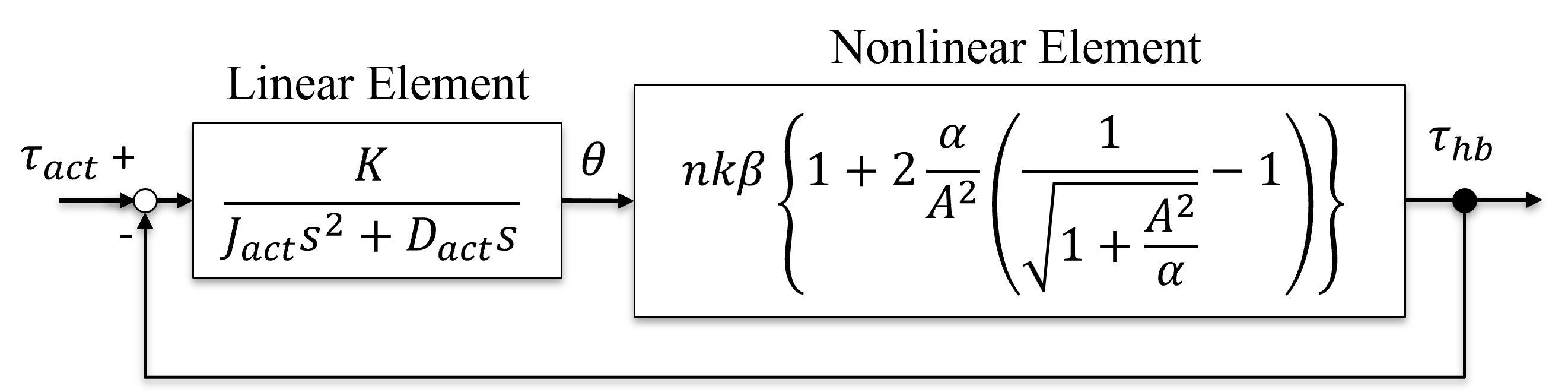}
    \caption{Feedback system representation of NSEA. }
    \label{fig:fbs}
\end{figure}

Hence, the actuator system, that incorporates the NSEE as an elastic element (as illustrated in Fig.\ref{fig:sea}), can be seen as a feedback system. This system comprises both linear and nonlinear elements, as depicted in Fig.\ref{fig:fbs}.

\section{APPLICABILITY OF DESCRIBING FUNCTION TO CONTROLLER}  \label{sec:applicability}

\subsection{Physical Simulation of NSEE}
The describing function, as discussed earlier, is a useful analytical tool. However, real-world validation is crucial to ensuring its applicability. Thus, we executed a physical simulation to confirm the results derived from the describing function in Sec. \ref{sec:dfnsee}.

\begin{figure}[http]
    \centering
    \includegraphics[width=205pt]{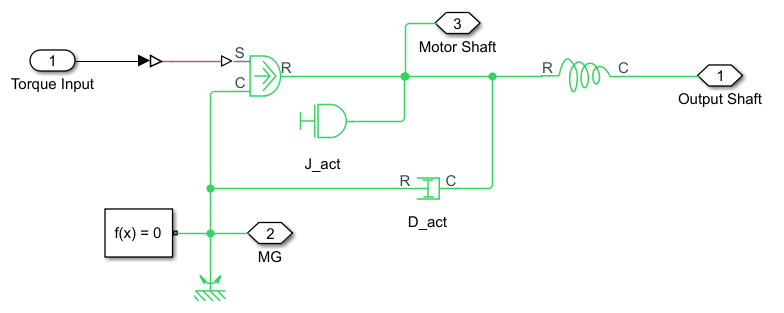}
    \caption{ Physical Simulation Model }
    \label{fig:phs}
\end{figure}

\begin{figure}[H]
    \centering
    \includegraphics[width=205pt]{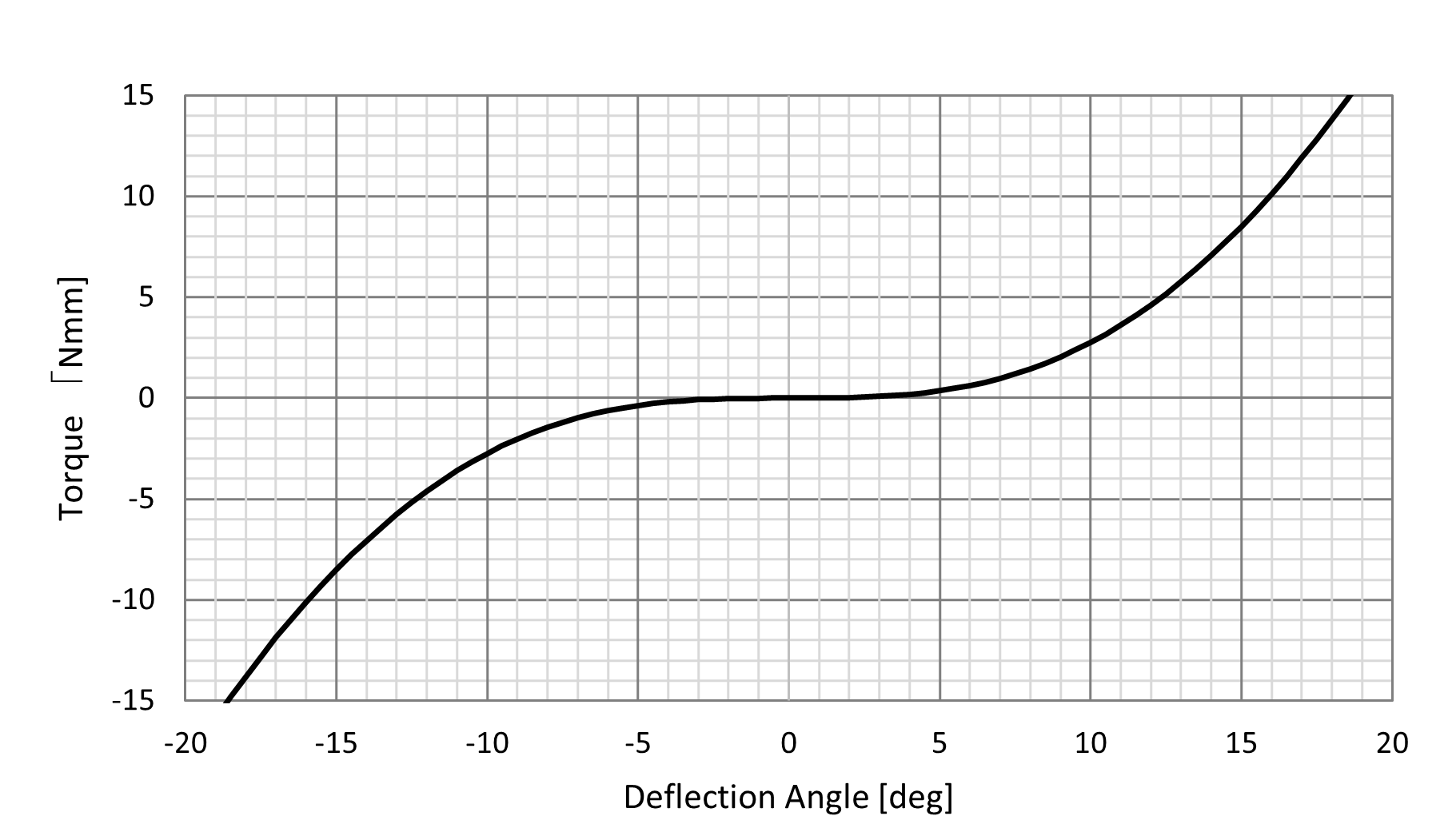}
    \caption{ Stiffness Characteristic of NSEE }
    \label{fig:nseechara}
\end{figure}

The physical simulation was conducted to verify the describing function we derived in Sec. \ref{sec:dfnsee} by using the Simscape library of MATLAB/Simulink environment.
Fig. \ref{fig:phs} shows the actuator physical model composed of an inertia element and a damping element, a nonlinear rotational spring element as NSEE based on Fig. \ref{fig:sea} and Table \ref{phypara}. Fig. \ref{fig:nseechara} shows the stiffness characteristic of NSEE in the model.
It was designed to achieve a saturation frequency of more than 15 Hz at torque amplitude of $ \pm 15 $ Nm from \eqref{eq:wsat} and \eqref{eq:kdes}, \eqref{eq:tns} and Table \ref{phypara}.
 The stiffness gradually increases with respect to the deflection angle. The nonlinearity makes compliance lower during stiffness is low, and control bandwidth higher during the stiffness is high than conventional SEAs.

\begin{figure}[H]
    \centering
    \subfloat[±15Nm, 2Hz]{\includegraphics[width=205pt]{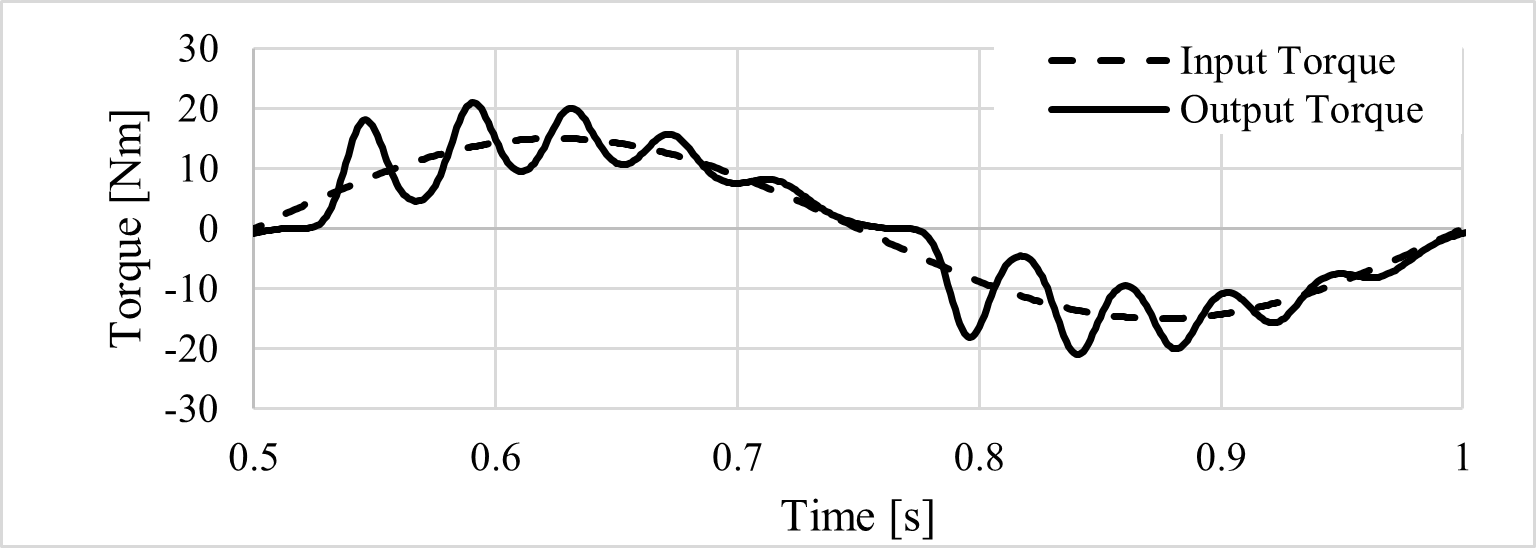}
    \label{fig:15Nm2Hz_phs}}
    \\
    \subfloat[±1Nm, 2Hz]{\includegraphics[width=205pt]{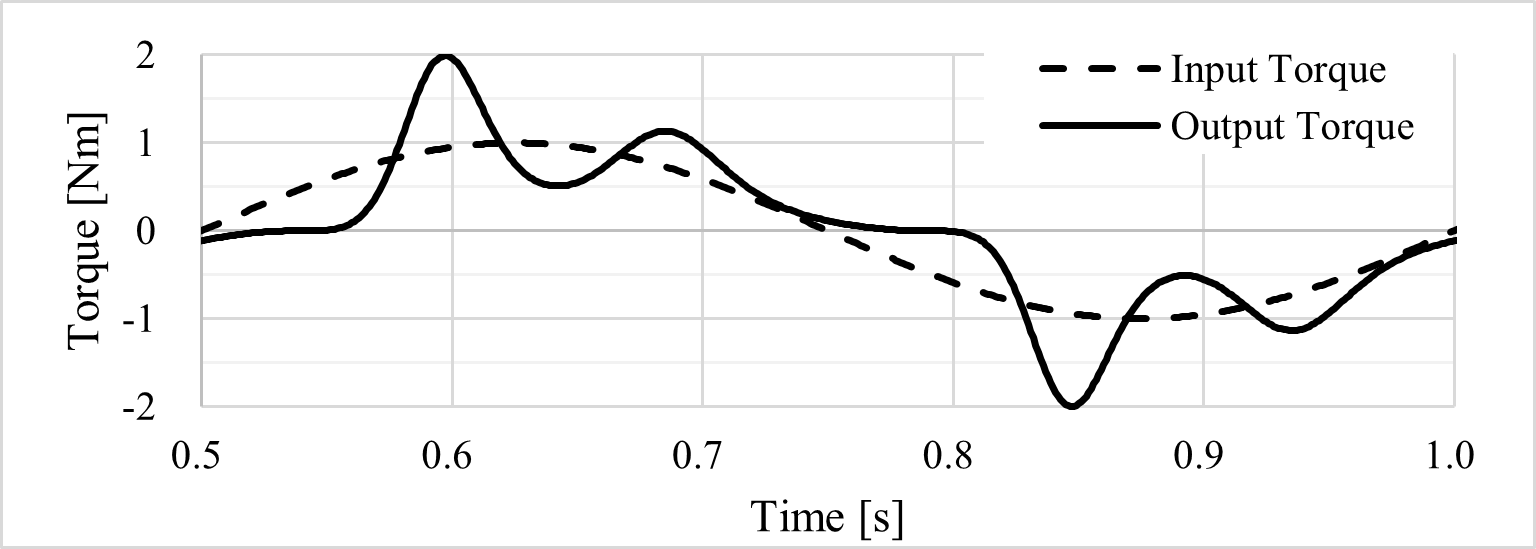}
    \label{fig:1Nm2Hz_phs}}
    \\
    \subfloat[±15Nm, 12Hz]{\includegraphics[width=205pt]{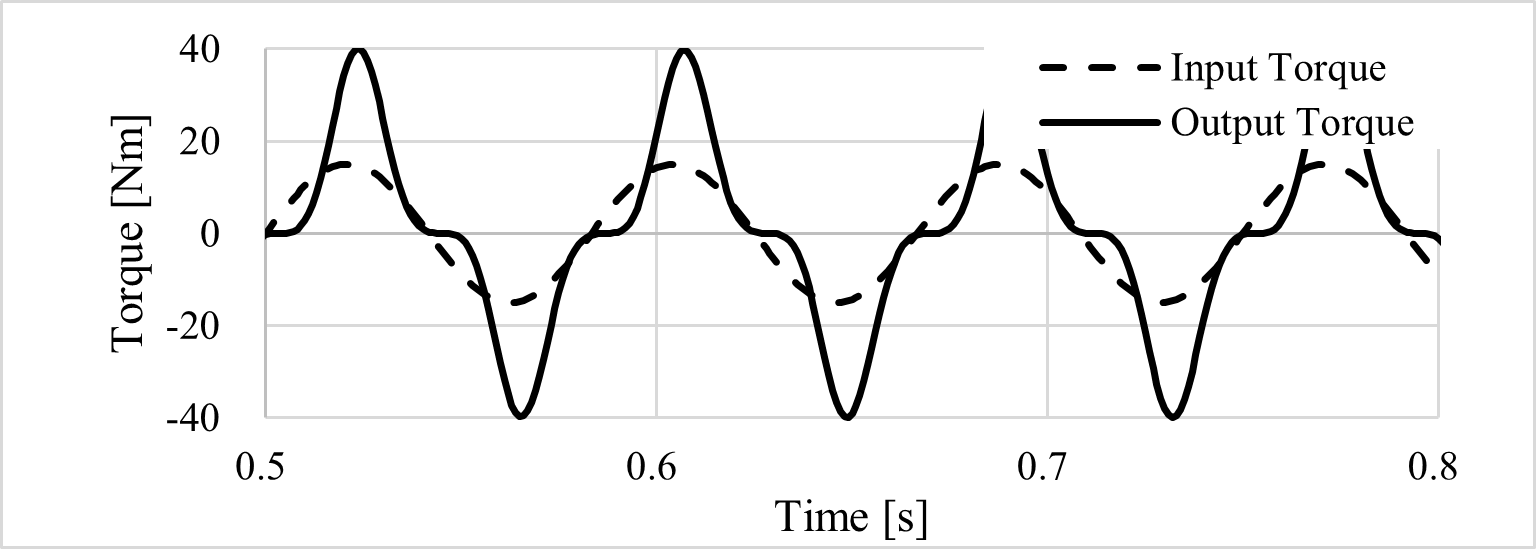}
    \label{fig:15Nm12Hz_phs}}
    \\
    \subfloat[±1Nm, 12Hz]{\includegraphics[width=205pt]{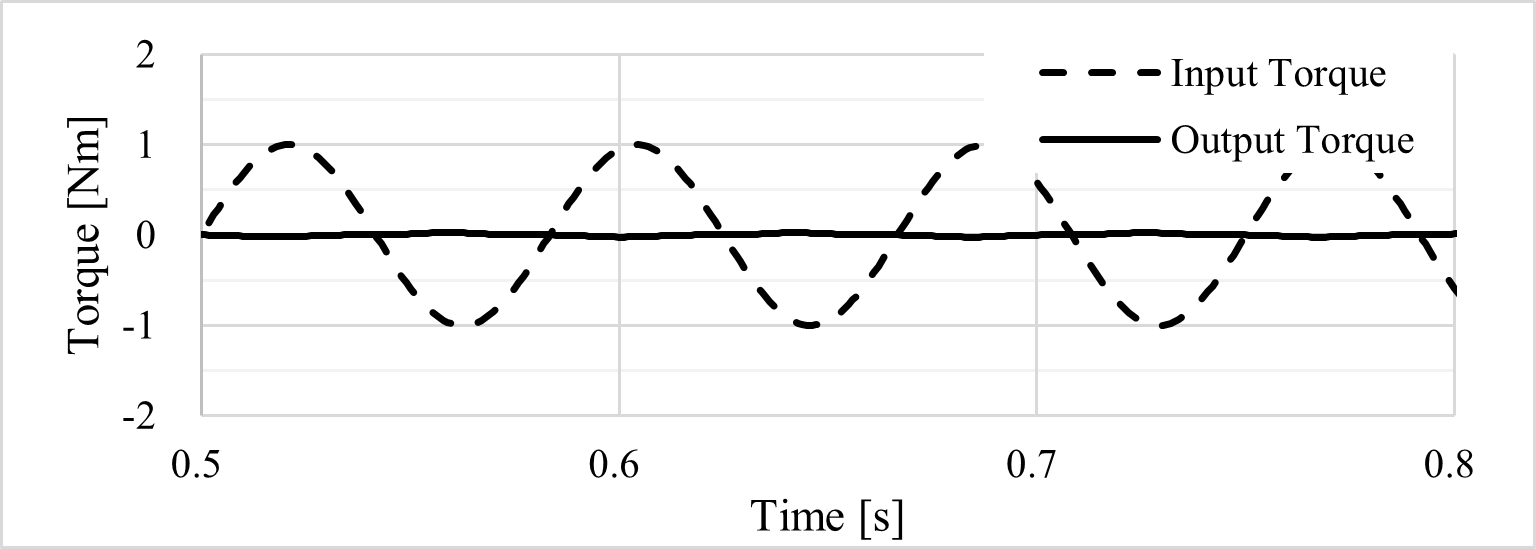}
    \label{fig:1Nm12Hz_phs}}
\caption{ Physical Simulation Results }
\label{fig:phsresult}
\end{figure}

Frequency response analysis was conducted by introducing sine wave inputs represented by $\tau_{act}$ and subsequently observing the output torque responses, $\tau_{hb}$. Fig. \ref{fig:phsresult} shows the simulation results with torque input of various amplitude and frequency.
Comparing Fig. \ref{fig:15Nm2Hz_phs} and Fig. \ref{fig:1Nm2Hz_phs}, higher frequency oscillation occurred with larger torque amplitude since the dynamics depend on the input amplitude by nonlinear stiffness as the describing function we derived in Sec. \ref{sec:dfnsee}.
In Fig. \ref{fig:15Nm12Hz_phs}, the output torque has a delay at the beginning of the period but still catches up at the peak of input torque. However, the actuator couldn't keep up with ±1 Nm input torque amplitude at the same frequency (Fig. \ref{fig:1Nm12Hz_phs}). 
It implies saturation torque change corresponding to the input torque amplitude. 
Since these results contain higher frequency oscillation than input, we computed the gain $G$ for frequency response analysis from $ \tau_{act} $ to $ \tau_{hb} $ by using RMS value as:
\begin{align} \label{eq:gain}
G = \frac{ \sqrt{ \int_{ T }^{ 2T } {\tau_{hb} }^2 (t) dt } }{ \sqrt{ \int_{ T }^{ 2T } {\tau_{act} }^2 (t) dt } }
\end{align}
where $T$ represents the period of the input sine wave.

\begin{figure}[http]
    \centering
    \vspace{0.1in}
    \includegraphics[width=205pt]{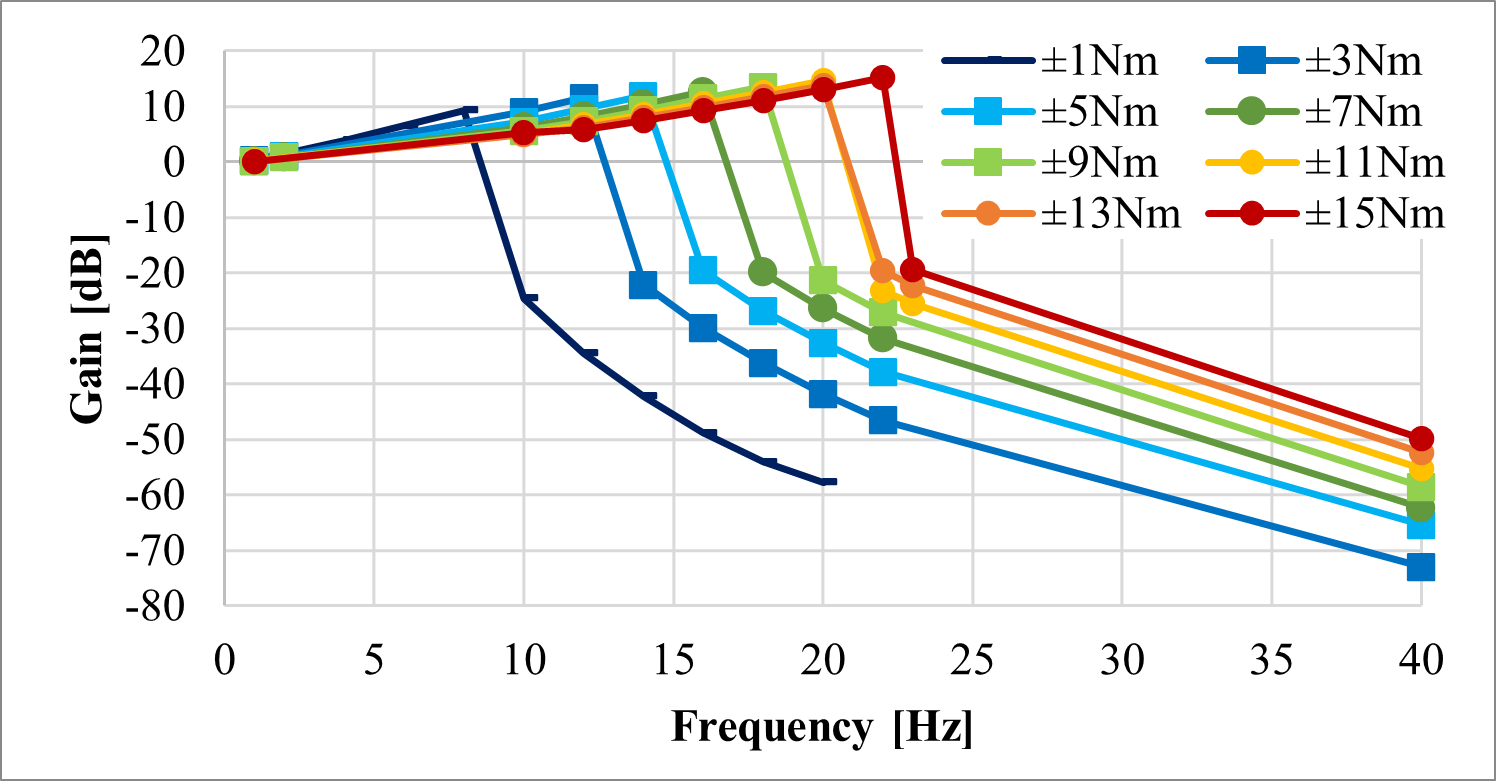}
    \caption{ Frequency Response Analysis Result by using Physical Simulation. }
    \label{fig:fa_phs}
\end{figure}

Fig. \ref{fig:fa_phs} shows the results of frequency response analysis by changing the amplitude of the input torque from ±1[Nm] to ±15[Nm]. As we mentioned, the saturation frequency varies through the torque amplitude. Note that the gains are immediately dropped below 0 dB after the natural frequencies, unlike linear stiffness.

\begin{table}[H]
\vspace{0.1in}
\caption{Actuator Parameters}
\label{phypara}
\begin{center}
\begin{tabular}{c|c|c}
\hline
Items & Units & Values \\
\hline
Actuator Inertia $J_{act}$ & $kgm^2$ & 0.005\\
\hline
Damping Coefficient $D_{act}$ & $Nm/(rad/s)$ & 0.1\\
\hline
Number of Tension Springs $n$ & - & 4\\
\hline
Stiffness of Tension Springs $k_{s}$ & $N/mm$ & 32\\
\hline
Length $R$ & $mm$ & 70\\
\hline
Length $r$ & $mm$ & 40\\
\hline
\end{tabular}
\end{center}
\end{table}

\subsection{Frequency Response Analysis with Describing Function}

We also conducted a frequency response analysis with The feedback system shown in Fig. \ref{fig:fbs}  by using MATLAB/Simulink. 
In the simulation, we need to define the input amplitude $A$ of describing function \eqref{eq:df} corresponding to input torque amplitude $A_{ \tau }$.  However, we can't derive directly $ \theta $ from torque $ \tau_{ns} $ by using \eqref{eq:tns}.
In the simulation, we sampled certain sets of $ (\theta, \tau_{ns} ( \theta ) ) $ in advance, and calculate $A$ from \eqref{eq:mac} by utilizing Newton-Raphson method~\cite{pho2022improvements}  as :

\begin{align} \label{eq:nr}
A = \theta_{i} - \frac{ \tau_{ns} ( \theta_{i} ) - A_{ \tau } }{ \tau_{ns}^{'} ( \theta_{i} ) } = \theta_{i} - \frac{ ( \tau_{ns} ( \theta_{i} ) - A_{ \tau } ) ( \alpha + \theta_i^2 )^2 }{ n k_s \beta ( 3 \alpha + \theta_i^2 ) \theta_i^2 } 
\end{align}

where $ \tau_{ns} ( \theta_{i} ) $ denotes the nearest $i$ th sampling torque to input torque amplitude $A_{ \tau }$.

\begin{figure}[H]
    \centering
    \vspace{0.1in}
    \subfloat[±15Nm, 2Hz]{\includegraphics[width=205pt]{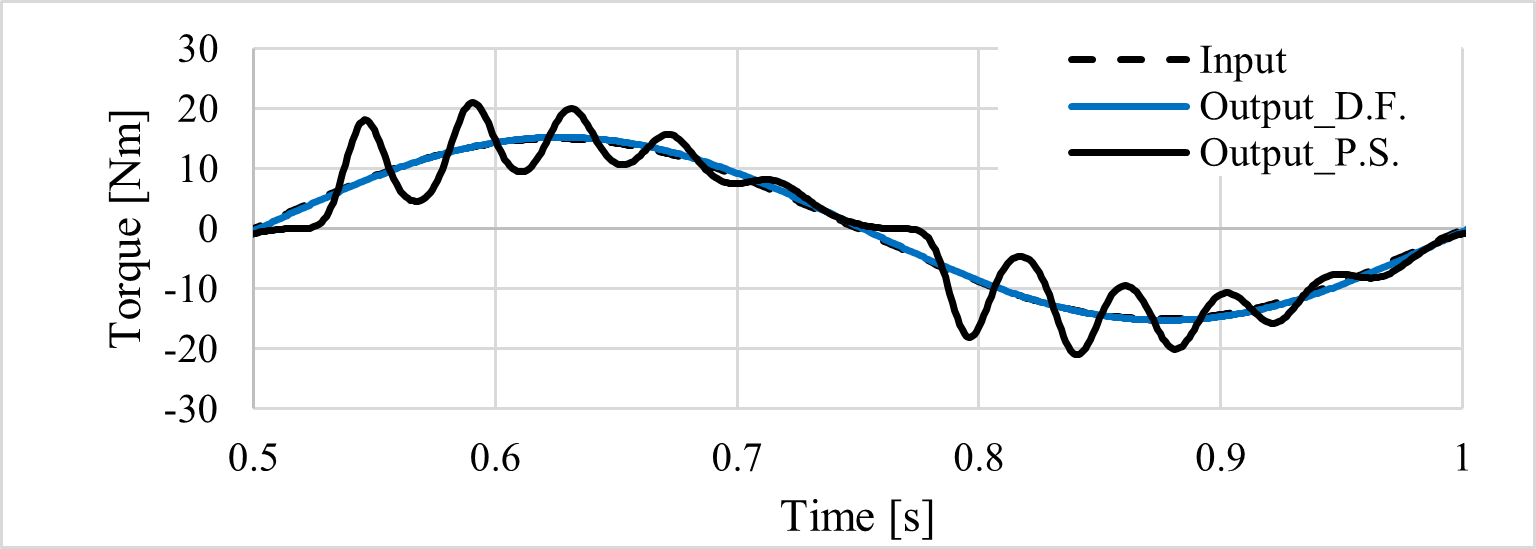}
    \label{fig:15Nm2Hz_dfs}}
    \\
    \subfloat[±1Nm, 2Hz]{\includegraphics[width=205pt]{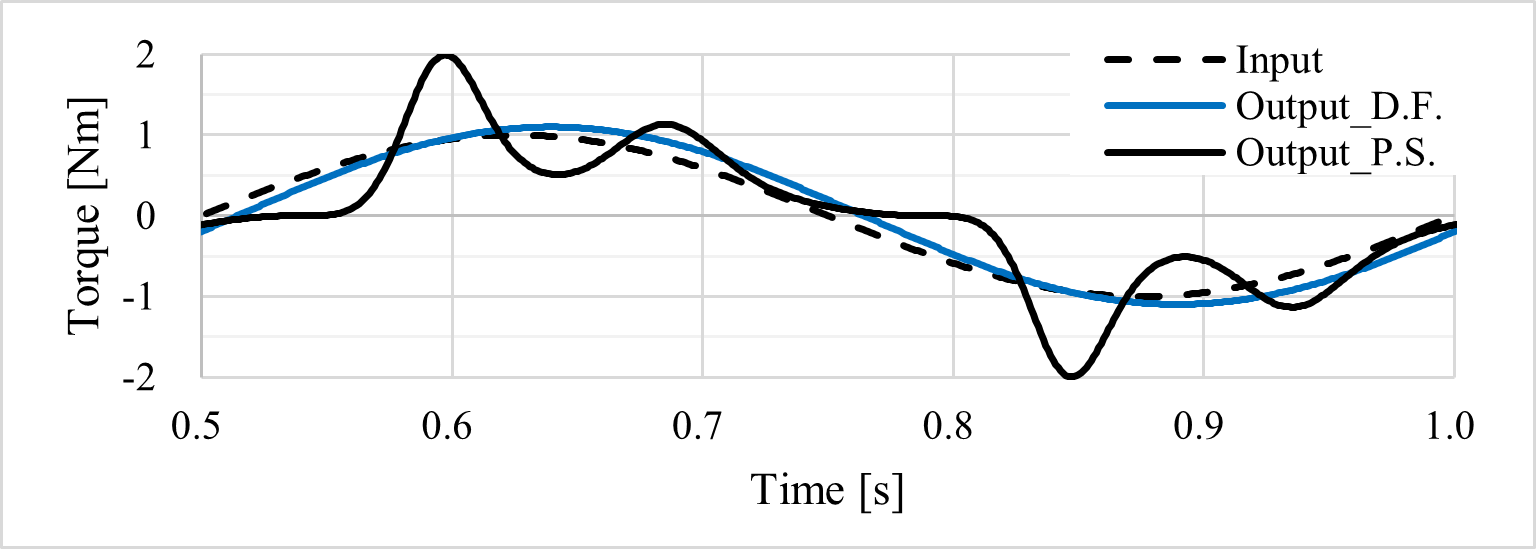}
    \label{fig:1Nm2Hz_dfs}}
    \\
    \subfloat[±15Nm, 12Hz]{\includegraphics[width=205pt]{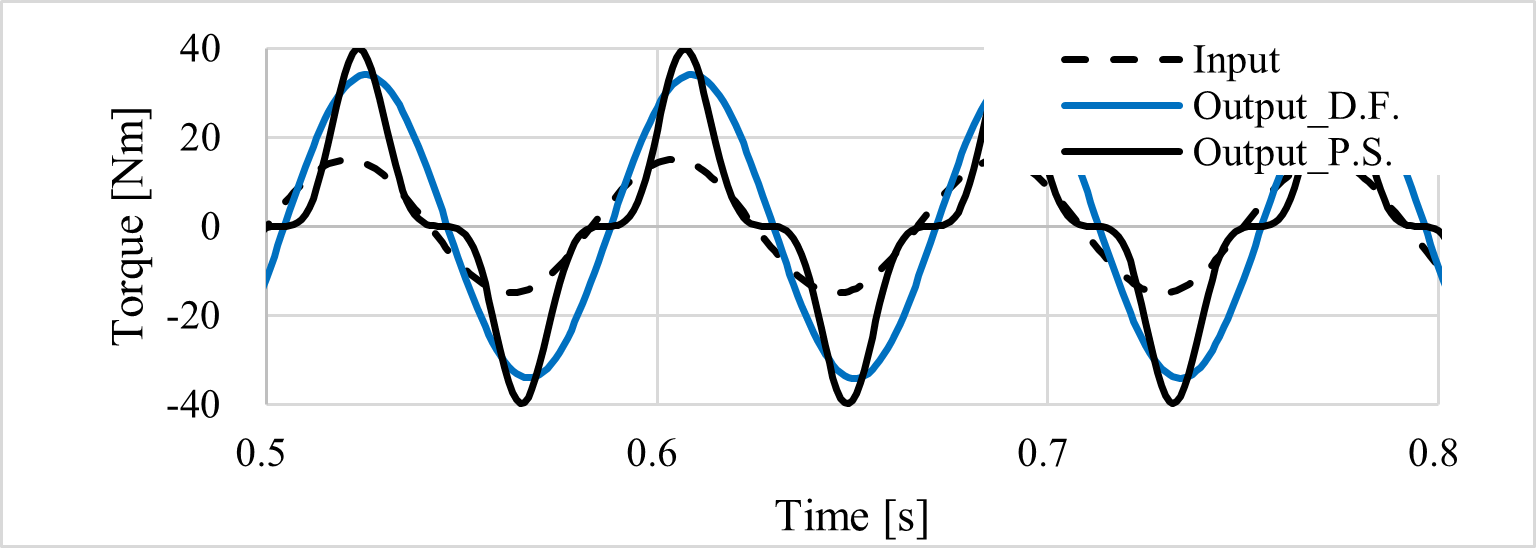}
    \label{fig:15Nm12Hz_dfs}}
    \\
    \subfloat[±1Nm, 12Hz]{\includegraphics[width=205pt]{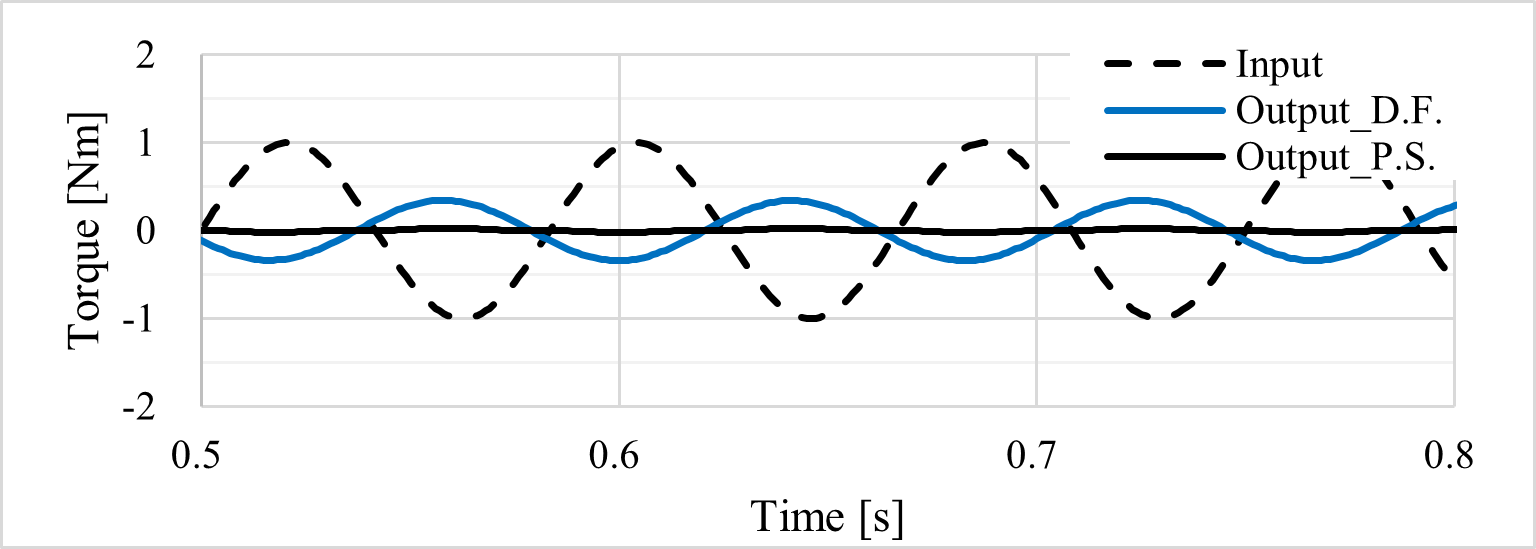}
    \label{fig:1Nm12Hz_dfs}}
\caption{ Simulation Results with Describing Function }
\label{fig:dfsresult}
\end{figure}

Fig. \ref{fig:dfsresult} compares the simulation results with torque input of various amplitude and frequency.
The simulation results are not completely matched especially when high-frequency oscillations occur, but still, the output of the describing function changes with following the center of the oscillation.

\begin{figure}[http]
    \centering
    \includegraphics[width=205pt]{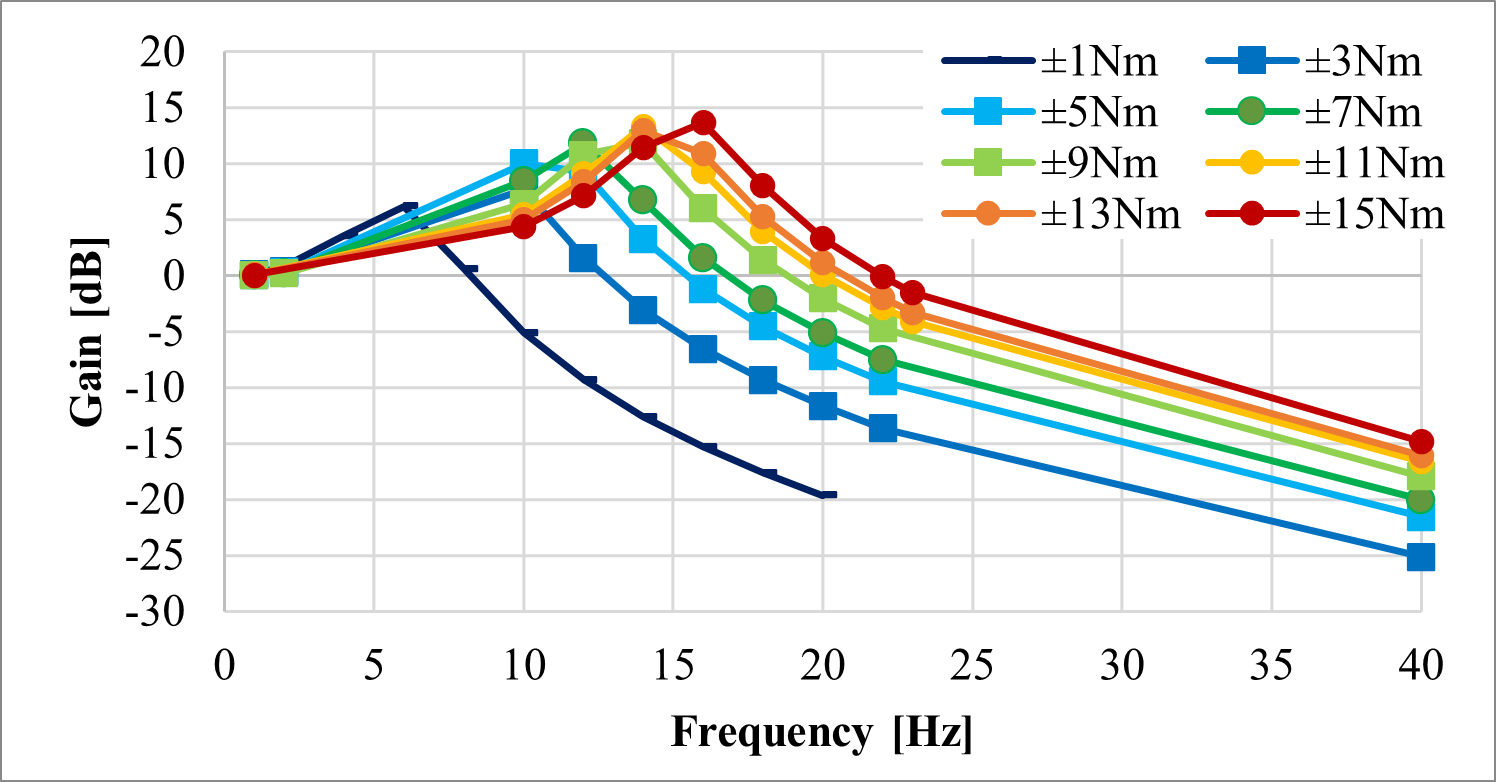}
    \caption{ Frequency Response Analysis Result by using Describing Function. }
    \label{fig:fa_dfs}
\end{figure}

Fig. \ref{fig:fa_dfs} shows the results of frequency response analysis by changing the amplitude of the input torque from ±1[Nm] to ±15[Nm] with describing function. Compared to Fig. \ref{fig:fa_phs}, the gain has no sudden drop after natural frequency because of approximation. Zero crossing frequencies between the physical model and the describing function model are compared in Table \ref{zcf}. Since the difference between these zero crossing frequencies is within ±1 Hz, the describing function derived in this paper is considered to capture the dynamic behaviors.

The describing function method offers valuable insights but presents certain limitations. It effectively reveals system behavior within a specific frequency range, however, its applicability to complex, high-dimensional systems and multivariable subsystems may be restricted. Additionally, its computational complexity can be substantial. The method is most effective when applied to systems with relatively constant input frequencies, thus, analyzing systems with highly dynamic input signals is challenging. Researchers should be mindful of these limitations and carefully consider their relevance when applying the describing function method to complex engineering systems.

\begin{table}[h]
\vspace{0.1in}
\centering
\caption{Zero Crossing Frequency Comparison between Physical Model and Describing Function [Hz]}
\label{zcf}
\begin{tabular}{lccccccccc}
\toprule
& \multicolumn{8}{c}{Input Torque Amplitude [Nm]} \\
\cmidrule(lr){2-9}
& ±1 & ±3 & ±5 & ±7 & ±9 & ±11 & ±13 & ±15 \\
\midrule
Physical \\ Model & 9 & 13 & 15 & 17 & 19 & 21 & 21 & 22 \\
\midrule
Describing \\ Function & 8 & 13 & 15 & 17 & 19 & 20 & 21 & 22 \\
\bottomrule
\end{tabular}
\end{table}

   

\subsection{Links to Controller Design of NSEA}

In the realm of NSEA control, gain scheduling controllers have been the traditional choice due to their less conservative nature~\cite{rafeedi2021design}~\cite{thorson2011nonlinear}. However, conventional gain scheduling controllers pose challenges in guaranteeing stability\cite{albayrak2022switching}. With the introduction of the describing function, a renewed approach to NSEA controller design emerges. From \eqref{eq:df} and Fig. \ref{fig:fbs}, NSEAs are modeled as a linear parameter varying (LPV) system:

\begin{align} 
\label{eq:lpv}
G  ( s, A ) = \frac{K N_{ \tau } ( A ) }{J_{act} s^2 + D_{act} s + K N_{ \tau } ( A )}
\end{align}

The LPV structure of \eqref{eq:lpv}, with coefficients as functions of the measurable parameter \(A\), makes it apt for $\mathcal{H}_\infty$ control design tailored for parameter variations~\cite{han2021robust}. Traditional $\mathcal{H}_\infty$ controllers tend to be conservative when addressing uncertainties or nonlinearities. The LPV nature of \eqref{eq:lpv} facilitates real-time adaptation to the parameter \(A\), ensuring optimal performance over its entire range.

Using the LPV formulation of \eqref{eq:lpv}, we can employ robust LPV-based $\mathcal{H}_\infty$ methodologies, leading to designs that are less conservative yet robust against the system's inherent nonlinearities and uncertainties. This model will be further formulated as a state feedback problem and embedded within a self-scheduled LPV control framework as outlined in \cite{sereni2020new}, \cite{apkarian1995self}.

\section{CONCLUSIONS}

This study presented the potential of the describing function method as a tool to analyze and understand nonlinear series elastic actuators, vital components in contemporary robotics. The describing function method is special because it helps design actuator controls and also reflects the qualities of natural muscles.

To validate our theoretical insights, we compared outcomes derived from the describing function with empirical results from physical models. The correlation supports the method's reliability and promise.

Nonlinear series elastic actuators, viewed as the robotic "muscles," dictate how robots maneuver and engage with their surroundings. Enhancing our understanding of these can pave the way for robots that emulate organic movement patterns. The describing function, with its muscle-like representation, emerges as a promising candidate for this purpose.

In essence, our exploration underscores the potential of the describing function method to recreate and understand the intricacies of muscle-like movement in robotics. 












\bibliographystyle{IEEEtran}
\bibliography{ROBIO2023}

\end{document}